%% file: main.tex
\documentclass[runningheads]{llncs}
\usepackage{graphicx}
\usepackage{subcaption}

\usepackage{tikz}
\usepackage{comment}
\usepackage{amsmath,amssymb} 
\usepackage{color}
\usepackage{graphicx}
\usepackage{booktabs}
\usepackage{bbm}
\usepackage{marvosym}

\usepackage[accsupp]{axessibility}  

\usepackage[pagebackref,breaklinks,colorlinks]{hyperref}

\usepackage[capitalize]{cleveref}
\crefname{section}{Sec.}{Secs.}
\Crefname{section}{Section}{Sections}
\Crefname{table}{Table}{Tables}
\crefname{table}{Tab.}{Tabs.}

\usepackage[width=122mm,left=12mm,paperwidth=146mm,height=193mm,top=12mm,paperheight=217mm]{geometry}

\begin{document}
\pagestyle{headings}
\mainmatter
\def\ECCVSubNumber{4814}  

\title{Contrastive Language-Action Pre-training for Temporal Localization} 

\titlerunning{ECCV-22 submission ID \ECCVSubNumber} 
\authorrunning{ECCV-22 submission ID \ECCVSubNumber} 
\author{Anonymous ECCV submission}
\institute{Paper ID \ECCVSubNumber}

\titlerunning{Abbreviated paper title}
%
\author{Mengmeng Xu\inst{1}\thanks{Work done during an internship with Amazon.}
\and
Erhan Gundogdu\inst{2}\Letter\thanks{Corresponding author.}
\and
Maksim Lapin\inst{2} \and \\
Bernard Ghanem\inst{1} \and
Michael Donoser\inst{2} \and
Loris Bazzani\inst{2}
}
\authorrunning{M. Xu et al.}
%
\institute{
King Abdullah University of Science and Technology\\
\{name.surname\}@kaust.edu.sa
\and
Amazon\\
\{eggundog, maksiml, donoserm, bazzanil\}@amazon.com
}
\maketitle
\input{sections/0_abstract.tex}
\input{sections/1_intro.tex}

\input{sections/2_related_work.tex}
\input{sections/3_method.tex}

\input{sections/4_experiments.tex}
\input{sections/5_conclusion.tex}

\clearpage
%
%
\bibliographystyle{splncs04}
\bibliography{egbib}
\end{document}

%% file: sections/0_abstract.tex
\begin{abstract}
Long-form video understanding requires designing approaches that are able to temporally localize activities or language.
End-to-end training for such tasks is limited by the compute device memory constraints and lack of temporal annotations at large-scale.
These limitations can be addressed by pre-training on large datasets of temporally trimmed videos supervised by class annotations.
Once the video encoder is pre-trained, it is common practice to freeze it during fine-tuning.
Therefore, the video encoder does not learn temporal boundaries and unseen classes, causing a domain gap with respect to the downstream tasks. 
Moreover, using temporally trimmed videos prevents to capture the relations between different action categories and the background context in a video clip which results in limited generalization capacity.
To address these limitations, we propose a novel post-pre-training approach without freezing the video encoder which leverages language.
We introduce a masked contrastive learning loss to capture visio-linguistic relations between activities, background video clips and language in the form of captions.
Our experiments show that the proposed approach improves the state-of-the-art on temporal action localization, few-shot temporal action localization, and video language grounding tasks.
\end{abstract}

%% file: sections/1_intro.tex
\section{Introduction}
\label{sec:introduction}

\begin{figure*}[t!]
    \centering
    \begin{subfigure}[b]{0.45\textwidth}
\includegraphics[width=.85\linewidth]{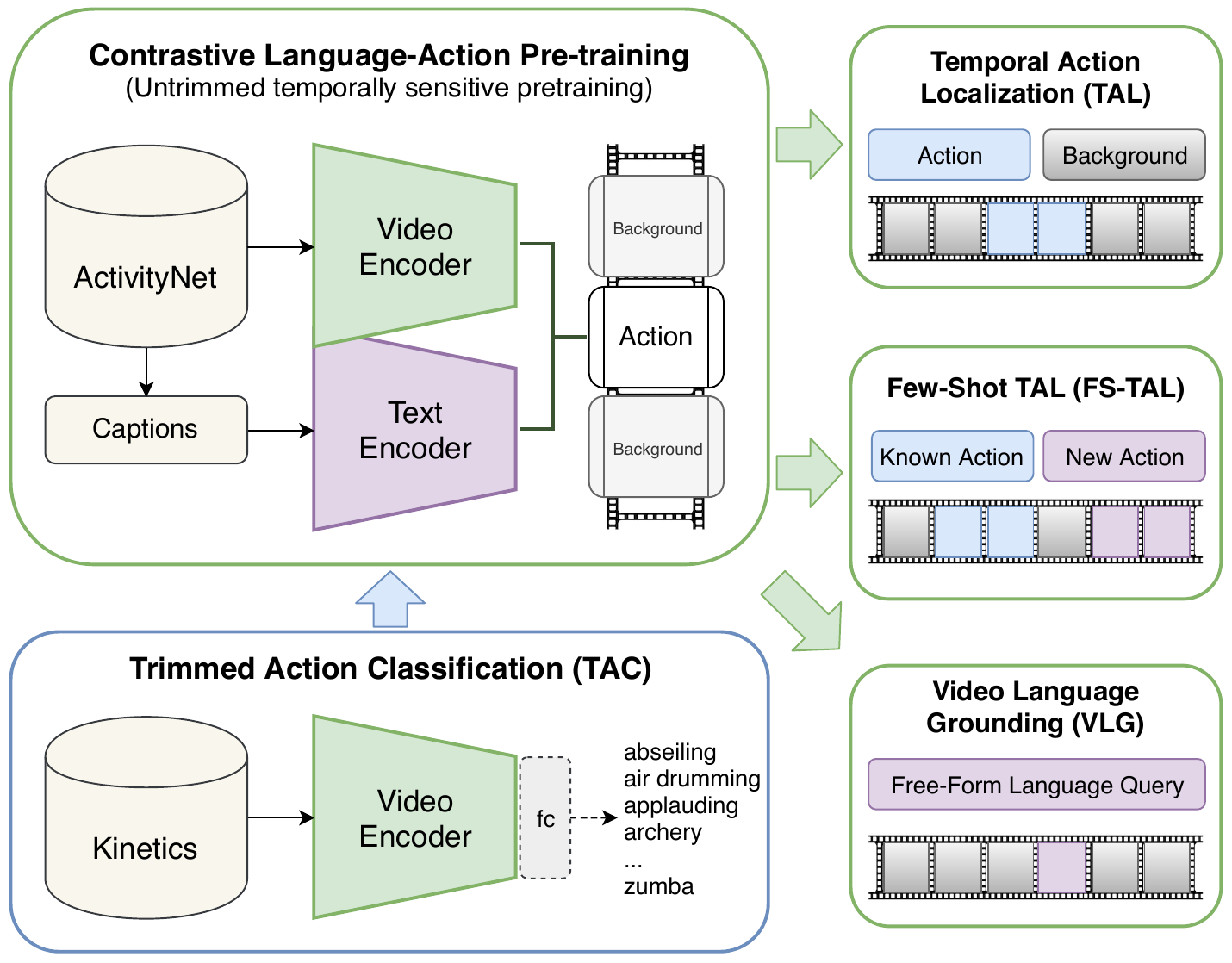}
\caption{}
\label{fig:clap-tasks}
    \end{subfigure}%
    ~
    \begin{subfigure}[b]{0.55\textwidth}
        \centering
\includegraphics[width=.99\linewidth]{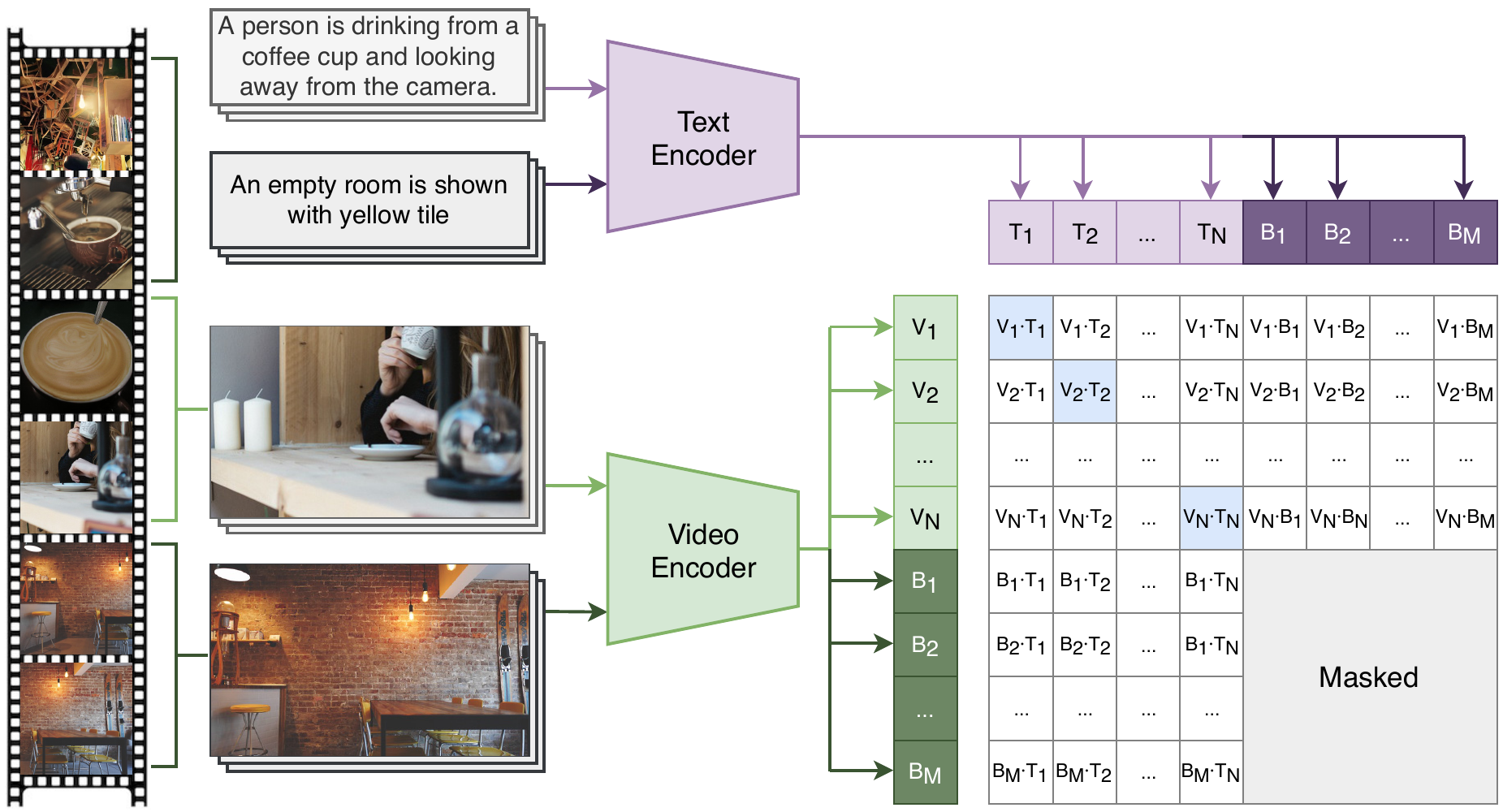}
\caption{}
\label{fig:clap-method}
    \end{subfigure}
\caption{
\textbf{(a) Contrastive Pre-training for Temporal Localization.}
Video encoders are pre-trained for trimmed action classification (TAC).
Then CLAP is a post-pre-training phase, where the video encoder is trained on relatively small datasets (no freezing) to capture visio-linguistic relationships between activities, background clips and language descriptions.
Finally, the video encoder can be used for downstream tasks: temporal action localization (TAL), few-shot TAL (FS-TAL), and video language grounding (VLG).
\textbf{(b) Contrastive Language-Action Pre-training (CLAP).}
Foreground clips, background clips and their respective language descriptions are extracted from the video.
Their representations are contrasted using our masked contrastive learning loss.
The background clips are masked and not contrasted to each other.
}
\end{figure*}


Video content widely available on services such as Netflix, YouTube, and Amazon Video is in long form.
One of the fundamental building blocks for long-form video understanding is temporal action localization.
Temporal localization in untrimmed long-form videos is challenging due to 
the technical constraints related to computational costs and the need to scale video understanding to previously unseen actions and more powerful video representations that can be driven by natural language.

Existing solutions for video understanding tasks, such as Temporal Action Localization (TAL)~\cite{xu2020g,Buch2017SSTST,caba2015activitynet} and video language grounding~\cite{Hendricks_2017_ICCV,Mun_2020_CVPR,gao2017tall,escorcia2019temporal}, are based on the paradigm of pre-training followed by fine-tuning~\cite{han2020memory,benaim2020speednet}.
A video encoding network is typically pre-trained on an action classification task using short trimmed video clips (e.g., Kinetics \cite{carreira2017quo}\footnote{Kinetics was created by trimming 10s clips around a frame manually labelled containing the target action. Thus the video does not contain background information.} or Sports-1M \cite{karpathy2014large}).
The encoding network is then frozen, applied as a feature extractor, and combined with additional layers that are fine-tuned on the downstream tasks.
The pre-training approach based on classification of short clips is sub-optimal for two reasons. 
1)~Each trimmed clip is used to optimize the video encoder independently, thus the encoder loses the capability to model the relation between multiple clips, which is especially important for temporal localization.
2)~The classification task limits the expressiveness of the learned representations because the encoder is not well suited for the videos that contain activity or language outside of the pre-defined set of classes.
These limitations create a gap between the pre-training process and the actual downstream tasks~\cite{alwassel_2021_tsp,alwassel_2018_detad},
which is approached in previous work by using strong supervision during fine-tuning.
 
Recent methods~\cite{miech20endtoend,xu2021videoclip,Luo2021CLIP4Clip} relax this assumption by proposing visio-linguistic contrastive learning to learn a universal representation that can be used for zero-shot learning or fine-tuned for downstream tasks.
Their main advantage is the ability to generalize beyond known action categories using language.
However, they require a large amount of data for pre-training and are not designed to improve temporal localization and grounding capability since there is no explicit notion of the \emph{background} context during the pre-training stage.

The main objective of this work is to propose a method that trains video encoders with language supervision and does not require large amount of training data like~\cite{miech20endtoend,xu2021videoclip,Luo2021CLIP4Clip}.
These video encoders are particularly suited for temporal action localization and video language grounding, because we contrast activities, background video clips and language during training using untrimmed videos. 
As a consequence, our method not only leads to superior results on TAL when compared to the state of the art, but also can be transferred to downstream tasks with unseen action categories or video language grounding.

In particular, the proposed approach, namely Contrastive Language-Action Pre-training (CLAP), leverages on classification pre-training for initializing a representation as depicted in Fig.~\ref{fig:clap-tasks}.
We adopt contrastive learning and extend it to language-activity pairs in a \textit{post-pre-training} phase.
This is possible by a combination of textual descriptions of video activities and synthetically generated language templates using activity labels.
The video encoder in CLAP is not frozen during post-pre-training and is trained end-to-end unlike previous contrastive learning approaches, \emph{e.g.}, VideoClip~\cite{xu2021videoclip}.
This gives our model the ability to adapt better and improve the video representation using relatively small datasets.
In addition, CLAP gains the ability to deal with long-form videos, since we can contrast multiple clips in each video against clips extracted from the other videos.
Finally, we introduce a loss function that masks out the video-language pairs that correspond to background, which is different from CLIP~\cite{radford2021learning} and VideoClip~\cite{xu2021videoclip}.
By omitting the contrastive loss for those pairs, we avoid pushing away the background segments in the feature space as they are semantically similar.
Our approach benefits from an action classification dataset (Kinetics-400~\cite{kay2017kinetics}) containing $400$k trimmed video clips and a dataset with temporal action annotations of untrimmed videos (ActivityNet~\cite{caba2015activitynet}) with $20$k videos unlike the contrastive video representation learning approaches \cite{miech20endtoend,xu2021videoclip,Luo2021CLIP4Clip} that leverage a large dataset \cite{miech19howto100m} containing $100$M text-video clip pairs and assume availability of significant computational resources.

Extensive experimentation shows the ability of CLAP to generalize on different downstream tasks. 
Our results for TAL on ActivityNet-1.3 show a significant improvement of $0.52\%$ $(\pm 0.07)$ in terms of average mAP compared to the state of the art~\cite{alwassel_2021_tsp}.
Moreover, CLAP outperforms previous methods on few-shot TAL with an improvement of $1.55\%$ avg.\ mAP compared to TSP~\cite{alwassel_2021_tsp} and $1.12\%$ avg.\ mAP to VideoCLIP~\cite{xu2021videoclip}.
On video language grounding, our method improves TSP~\cite{alwassel_2021_tsp} on Charades-STA by $2.4\%$ and on TACoS~\cite{TACoS_ACL_2013} by $0.8\%$ in mIoU.
Finally, our ablation studies show consistent improvements of CLAP on all tasks.


%% file: sections/2_related_work.tex
\section{Related Work}
\label{sec:related_work}


\paragraph{Pre-training for temporal localization.}
State-of-the-art approaches for temporal localization tasks adopt pre-training to learn video encoders 
in a supervised way for \textit{classification} on large-scale datasets, e.g., Kinetics~\cite{zisserman2017kinetics,Zeng_2020_CVPR}. 
In most TAL methods~\cite{lin2018bsn,lin2019bmn,xu2020g,Bai2020bcgnn}, features are extracted with TSN~\cite{wang2016temporal,trn_eccv18}, two-stream I3D~\cite{carreira2017quo}, e.g., \cite{he2019rethinking,Zeng2019GraphCN}, Pseudo-3D~\cite{qiu2017learning}, e.g., \cite{Long2019GaussianTA}.
Following this trend, proposal-based~\cite{chenhierarchical,ghosh_etal_2019_excl,Liu_2018_ECCV,GDP_2020_AAAI,lu_etal_2019_debug,chen2020learning,yuan2019to,zhang_etal_2020_span}, ranking~\cite{Ge_2019_WACV,song2018val,chen_etal_2018_temporally,wang2020temporally,YuDLZ00L18,zhang2019man}, and boundary regression methods~\cite{Hendricks_2017_ICCV,liu2018crossmodal,wu2018multi,Mun_2020_CVPR} for video language grounding also use pre-trained video encoders.
Since pre-training on large-scale datasets brings data gap to the target domain, some methods train the video encoder via temporal segment annotations from the downstream datasets by action classification~\cite{lin2019bmn,xu2020g,Rodriguez_2020_WACV,Mun_2020_CVPR},
or action localization~\cite{xu2017r,liu2020progressive}. However, the training targets in these methods are either discrepant from localization or challenging to achieve. 

Self-supervised way is an alternative to sidestep end-to-end training of video encoders~ \cite{alwassel2020xdc,benaim2020speednet,miech20endtoend,shuffle_learn_eccv16,arrow_cvpr18}.
Some works use temporal-related pretext tasks, such as frame ordering, either learning through triplets of frames \cite{shuffle_learn_eccv16}, through sorting a sequence \cite{sorting_seq_iccv17} or by distinguishing whether sequences are played forward or backward \cite{arrow_cvpr18}.
Other pretext tasks are related to the speed of the video~\cite{benaim2020speednet,playback_rate_cvpr20,wang2020self,temporal_transform_eccv20}, predicting motion-related statistics~\cite{stats_cvpr19} and more direct extensions of successful image-based contrastive learning methods to the video realm~\cite{st_contrastive_arxiv20}.

To summarize, most (self-)supervised pre-training methods require fine-tuning to close the gap between the downstream task and the pre-training task.
Video encoders are often used as fixed feature extractors for the downstream tasks,
and each clip is independently considered during optimization of the encoder.
CLAP overcomes these issues by 
1)~post-pre-training to reduce the gap without the need of large datasets for the downstream tasks; 
2)~unfreezing the video encoder, resulting in a model that can better adapt to the nuances of the post-pre-training task;
3)~gaining the ability to learn relationships between multiple clips from the same video and actions from other videos via contrastive learning.

\paragraph{Language-Video Pre-training.}
The availability of large image-text datasets collected from the Web and computational power made the pre-training of models for image-language alignment possible, such as CLIP~\cite{radford2021learning} and ALIGN~\cite{jia_2021_ICML_ALIGN}. 
More recently, a similar trend is showing up in the video understanding domain.
Cross-modal encoders concatenate the language and video representations and optimize modality-specific and cross-modality masking losses~\cite{sun_2019_CVPR_videobert,li_2020_AAAI_unicoder,zhu_2020_CVPR_actbert}.
Video-language contrastive learning compares positive video clip-text pairs with hard negatives~\cite{xu2021videoclip,Luo2021CLIP4Clip}.
VideoCLIP~\cite{xu2021videoclip} extends CLIP to videos by pre-training with temporally overlapped pairs of
video and text clips and selection of negatives via clustering to form batches. 
CLIP4Clip~\cite{Luo2021CLIP4Clip} transfers the knowledge from a pre-trained CLIP model to videos by projecting visual temporal tokens with 3D convolutions.
Such approaches consider video clips holistically and train universal models that can be applied to as many downstream tasks as possible.
In contrast, CLAP is trained to be action-aware via prompt engineering of language templates and using natural language descriptions which enable the video encoder to adapt better to the action-focused downstream tasks.
In addition, the video  backbone of CLAP is not frozen, thus enabling it to be more flexible in learning relationships between clips in the contrastive learning setup.

\paragraph{Post-pre-training.}
The pre-training of video models requires large computational resources.
Therefore, pre-trained models are used to extract features from clips and additional models are trained on top of those to improve performance on specific downstream tasks.
The data and task gap between the pre-training and finetuning has motivated previous works~\cite{LOFITAL,alwassel_2021_tsp}  to add a second step of pre-training before finetuning, namely \emph{post-pre-training}.
The objective of post-pre-training is to extend and tailor the model to the nuances of the downstream tasks, without using the strong supervision of those tasks.
LoFi~\cite{LOFITAL} performs post-pre-training by 
reducing the mini-batch composition so that jointly optimizing the video encoder and
TAL head becomes operable.
TSP~\cite{alwassel_2021_tsp} adopts a classification loss to distinguish video clips of different actions and background in a TAL dataset. 
Language-video pre-trained models VideoCLIP \cite{xu2021videoclip} and CLIP4Clip~\cite{Luo2021CLIP4Clip} use a form of post-pre-training, by training models on top of extracted features, e.g. via stopgrad, where they leverage on large-scale datasets (Howto100M and JFT-300M).
Unlike~\cite{xu2021videoclip,Luo2021CLIP4Clip}, CLAP introduces a post-pre-training method that does not require a large-scale dataset with temporal annotations, enabling to unfreeze the backbone similar to~\cite{alwassel_2021_tsp,LOFITAL}. 
Moreover, CLAP gets closer to the localization downstream tasks by prompt engineering of language templates with actions. 


%% file: sections/3_method.tex
\section{Method}
\label{sec:method}


The two stage paradigm of video encoder pre-training on action classification 
followed by fine-tuning of the localization head with the fixed encoder 
creates a gap in training and leads to suboptimal performance on the downstream task~\cite{alwassel_2021_tsp,alwassel_2018_detad}.
Features obtained from the video encoder trained on trimmed action classification
are not sensitive to action boundaries since the background context is not observed
during the video encoder training.
Instead of relying entirely on the fine-tuning step to close that gap, 
we introduce CLAP and describe a contrastive loss formulation
where activity clips are considered jointly with background video segments allowing
the model to learn sharp action boundaries, 
which is crucial for temporal action localization and generalization to unseen categories,
and language-related downstream tasks such as video language grounding.

\paragraph{Language-Action Alignment.}
Given an untrimmed video, we train a video encoder to distinguish the semantics
of different actions and their background context as showed in Figure~\ref{fig:clap-method}.
We use textual language to describe activities and leverage an off-the-shelf language model as a text encoder.
Video clips are contrasted with textual descriptions of actions (background)
in a semantic space, which is learned jointly end-to-end.
While our approach is agnostic to the specific form of textual descriptions and 
can learn from natural language supervision, 
we describe a simple template to generate
descriptions from a fixed set of classes and 
optionally from the available captions
in ActivityNet~\cite{caba2015activitynet}.
Experiments show that video clip and text alignment allows CLAP to exploit 
knowledge transfer from the pre-trained video and text encoders 
to build a better semantic space compared to prior work~\cite{alwassel_2021_tsp,Xu_2021_ICCV,LOFITAL},
generalize to unseen activity classes in a few-shot setting,
and achieve strong performance in video language grounding.

\subsection{Video and Language Encoders}

We use R(2+1)D~\cite{tran2018closer} pre-trained on Kinetics-400~\cite{kay2017kinetics}
as a video encoder and train it end-to-end on video clips
of size $3 \times L \times H \times W$,
where $L$ is the number of frames, and $H$ and $W$ are the frame height and width, respectively.
Video clips and textual descriptions are respectively denoted by $x_v$ and $x_t$.
We feed a video clip $x_v$ through the video encoder $f_{\theta_v}$ followed by 
a projection module $g_{\gamma_v}$ to obtain an embedding $z_v$ in the language-video space as
\begin{equation}
\label{eq:vid_rep}
z_v = g_{\gamma_v}\left(f_{\theta_v} \left(x_v \right) \right) ,
\end{equation}
where $\theta_v$ and $\gamma_v$ are the learned parameters of the video backbone and the projection module respectively. 
For downstream tasks, we drop the projection and use the representation directly 
from the video encoder $h_v =f_{\theta_v} \left(x_v \right)$ with fixed parameters $\theta_v$.

Similar to video clips, we obtain language features $z_t$ from a pre-trained language model $f_{\theta_t}$
(we use BERT~\cite{BERTPretrained} as a text encoder), followed by a learned projection module $g_{\gamma_t}$ 
that maps the features to the language-video embedding space as
\begin{equation}
\label{eq:text_rep}
z_t = g_{\gamma_t}\left(f_{\theta_t} \left(x_t \right) \right) ,
\end{equation}
where $\theta_t$ are the fixed parameters of the language model backbone,
and $\gamma_t$ are the learned projection weights.
Our projection modules $g_{\gamma_v}$ and $g_{\gamma_t}$ are simple feed-forward networks.

\subsection{Textual Language Generation}

We propose to represent background and action clips with language in different ways.
In the case that the dataset does not contain expressive descriptions, we use the action category and generate short prompts in the form of \emph{``background of \textless action\textgreater"} for background clips and \emph{``foreground of \textless action\textgreater"} for foreground. 
In contrast with the typical one-hot encoding, this prompts explicitly define the relation of background and foreground with respect to the action. 
We found this to be practically useful in contrast to one-hot encoding, and it leads to better performance.
Note that there exist other more complex ways to encode relationships between a clip in a video with language, e.g., encoding the relative temporal order of actions. 
We leave the exploration of relationship encoding for future work.

We introduce a second way to leverage on dense video captions available from public datasets (Activity-Net captions in this work) and make an assignment from dense video captions to sampled foreground and background action clips. 
Concretely, we assign a clip to its temporally closest caption with a non-zero overlap (if available). If there is no overlap, we use synthetically generated prompts. 
Available captions in the public datasets generally do not describe the background clips. Therefore, we propose a masked loss function which will be introduced in the next section. During training, we have $0.5$ probability to use captions and synthetically generated prompts.

In the rest of the paper, we use the terms synthetically generated prompts and language templates, interchangeably.

\subsection{Language-Action Contrastive Learning}
We train the video encoder by optimizing a contrastive loss~\cite{InfoNCE,chen2020simple} similarly to CLIP~\cite{radford2021learning}. 
We first introduce the original loss as follows:
\begin{equation}
\label{eq:clip_loss}
\mathcal{L}_{clip} = \sum\limits_{(v,t)\in B} \log \mathrm{NCE}(z_v, z_t) + \log \mathrm{NCE}(z_t, z_v)
\end{equation}
where $B$ is the set of positive pairs of video clip and text descriptions, and
$\mathrm{NCE}(z_v,z_t)$ is the contrastive loss defined as follows:
\begin{equation}
\label{eq:NCE_loss}
\mathrm{NCE}(z_v, z_t) = \frac{\exp(z_v \cdot z_t / \tau)}{\sum\limits_{z \in \{ z_t , z_t^- \} } \exp(z_v \cdot z / \tau)}
\end{equation}
where $z_v \cdot z_t$ is the dot product, $\tau > 0$ is the temperature parameter,
and $\{z_t^-\}$ is the set of negative text descriptions for $z_v$ in the batch.
$\mathrm{NCE}(z_t, z_v)$ is defined in the same way, but we swap the indices $t$ and $v$ in Eq.~\eqref{eq:NCE_loss}.
MIL-NCE~\cite{miech20endtoend} and VideoCLIP~\cite{xu2021videoclip} introduce variants of Eq.~\eqref{eq:clip_loss} to deal with action recognition and text-to-video retrieval tasks.
They pre-train on datasets with the noisy labels (HowTo100M~\cite{miech19howto100m}) and focus on negative and positive sampling.
Our goal is different from these approaches as we design pre-training to specifically deal with localization tasks on a smaller but cleaner dataset, by contrasting foreground and background clips with language.

The loss in Eq.~\eqref{eq:clip_loss} is a strong baseline for video encoder pre-training,
as we show in experiments in Sec.~\ref{sec:experiments}.
However, Eq.~\eqref{eq:clip_loss} has a critical limitation that leads to suboptimal performance
in our setting which we address next.
As Fig.~\ref{fig:clap-method} depicts at the background segments $B_i$, the original loss aims to model each background clip and its textual description as a separate class,
similar to the foreground actions.
This adds unnecessary complexity and forces the model to spend capacity on modeling the fine-grained background context, which is not needed in our downstream applications.
Moreover, the captions assigned to background clips are not as descriptive as for the foreground clips because available textual descriptions are more specific to defining the foreground activities.


To address the problem described above, we propose the following masked contrastive loss:
\begin{equation}
\label{eq:proposed_loss}
\begin{aligned}
\mathcal{L}_{mask}=\sum\limits_{(v,t)\in B} \log \mathrm{NCE}(z_v, z_t) \mathbbm{1}\left[ v \in \mathrm{F_g} \right] +\\
 \log \mathrm{NCE}(z_t, z_v) \mathbbm{1}\left[ t \in \mathrm{F_g} \right]
\end{aligned}
\end{equation}
where $\mathrm{F_g}$ is the set of foreground indices $1, \ldots, N$ of video and text descriptions,
and $\mathbbm{1[\cdot]}$ is the indicator function that outputs $1$ if the argument is true and $0$ otherwise.
In other words, we use the loss in Eq.~\eqref{eq:clip_loss} for the foreground text and video clips only, while maintaining all the background segments as negatives in Eq.~\eqref{eq:NCE_loss}.
By omitting the positive contrastive loss for background clip and text, the constraint in the feature space to push away the semantically similar backgrounds is relaxed, 
in turn, helping to improve the downstream tasks.


\subsection{Final Loss Function}
In addition to the proposed loss in Eq.~\eqref{eq:proposed_loss}, we also adopt the classification loss as in TSP~\cite{alwassel_2021_tsp} because it has been shown be effective for TAL. 
The classification loss is defined as follows:
\begin{equation}
\label{eq:class}
\mathcal{L}_{ce} = \left\{
        \begin{array}{ll}
             \mathcal{L}_{ce}(\widehat{y^r}, y^r) + \mathcal{L}_{ce}(\widehat{y^c}, y^c) & \quad if \quad y^r=1 \\
             \mathcal{L}_{ce}(\widehat{y^r}, y^r) & \quad otherwise
        \end{array}
    \right.
\end{equation}
where $\mathcal{L}_{ce}$ is the cross entropy loss over the class logits $\widehat{y^c}=f_{\theta_c}(f_{\theta_v} \left(x_v \right))$ 
with $f_{\theta_c}$, a projection layer to the $C$ action classes, and similarly $\widehat{y^r}=f_{\theta_r}(f_{\theta_v} \left(x_v \right))$ is a two dimensional projection layer to output two class logits for background and foreground.

Our final loss is the summation of Eq.~\eqref{eq:class} and Eq.~\eqref{eq:proposed_loss} as: $\mathcal{L}_{ce} + \mathcal{L}_{mask}$, named as CLAP. It employs both language prompts synthetically generated from actions and dense captions. 




%% file: sections/4_experiments.tex
\section{Experiments}
\label{sec:experiments}
We experimentally validate our approach in terms of following properties: 
1) efficacy of our video encoder on TAL (Sec.~\ref{sec:TAL_res}), 
2) generalization capability of our approach to both unseen categories in a few-shot TAL setting (Sec.~\ref{sec:few_shot_TAL_res}), and
3) performance on tasks that require language-video understanding (Sec.~\ref{sec:temporal_video_caption_grounding}). 
In addition, we perform ablation studies to compare the proposed loss and its variants to show the impact of background masking in the loss function and additional natural language descriptions from dense captions.
\subsection{Experiment setup}
\label{sec:experimental_setup}

\paragraph{Datasets.}
We employ three popular video datasets to evaluate the performance.

\noindent {\em (1) ActivityNet-1.3}~\cite{caba2015activitynet}  
is a benchmark for temporal action localization.
%
It contains 19,994 annotated untrimmed videos with 200 different action classes.
The split ratio of train:validation:test
is 2:1:1. 
Following the common practice, we train and test the models on the training and validation set. 
We adopt this dataset for TAL and few-shot TAL tasks.

\noindent{\em (2) Charades-STA}~\cite{gao2017tall} 
is a video language grounding dataset extended from 
the action recognition dataset Charades~\cite{sigurdsson2016hollywood}.
It contains $9,848$ videos of daily indoor activities, with $12,408$/$3,720$ moment-sentence pairs in train/test set respectively.
Language descriptions are shorter than in ActivityNet: the average number of words per caption is 6.23 for Charades-STA and 13.58 for ActivityNet.

\noindent{\em (3) TACoS}~\cite{TACoS_ACL_2013} consists of videos selected from the MPII Cooking Composite Activities video corpus~\cite{mpii}. 
It comprises 18,818 video-query pairs of different cooking activities. 
Each video contains an average of 148 queries, some of which are annotations of short video segments. Since TACoS dataset mainly has cooking videos, there is a bigger domain gap between TACoS and ActivityNet clips. 

\input{tables/TAL_main.tex}


\paragraph{Video clip sampling.}
We follow the video pre-processing setup of TSP~\cite{alwassel_2021_tsp}. 
In particular, 
all videos are re-encoded with a fixed frame rate of $30$ fps.
$16$-frame video clips are formed to cover approximately one second duration by uniform sampling.
We resize frames with the small dimension of $128$ pixels and apply random $112\times 112$ cropping during training and center cropping during testing.
Temporally contiguous segments are formed from videos with foreground (action) and background (no action) labels. 
We select 5 clips per segment and randomly and uniformly select them during training and testing, respectively.

\paragraph{Video and text encoders.}
We use R(2+1)D~\cite{tran2018closer} with ResNet-18 to encode video clips. 
Following the common practice in TAL, the R(2+1)D model is pre-trained on Kinetics-400.
We discard the last fully-connected layer to output a 512 dimensional vector for each video clip.
We represent the language description for each clip by a pre-trained BERT model~\cite{BERTPretrained}.
The representation is from the hidden layer of 768-dimensional \textsc{CLS} token.
We project the video and text encodings to a new space with the same 512 dimensionality using layers that consist of a batch normalization followed by a linear layer. 
We also report results using Res3D-18 on TAL.

\paragraph{Post-pre-training details.}
Training runs on 4 NVIDIA V100 GPUs with 16GB of memory.
Our method is implemented in PyTorch 1.10 with CUDA 10.2. 
We use distributed SGD to optimize our video backbone, classification head, and contrastive learning head. 
The learning rate is $10^{-4}$ for backbone encoder, $2\times 10^{-2}$ for linear layers in the other two heads. 
We keep the language model frozen during training. 
We train for $8$ epochs, and decrease the learning rate by a factor of $\gamma=0.01$ after every $2$ epochs.

\subsection{Efficacy on Temporal Action Localization}
\label{sec:TAL_res}

TAL aims to predict the boundaries of a predefined set of actions in untrimmed videos and recognizing the detected foreground activity.
We use mean Average Precision (mAP) at varying temporal Intersection over Union (tIoU) thresholds to evaluate the models.
Following the official evaluation setting, we report mAP scores at three tIoU thresholds of $\{0.5, 0.75, 0.95\}$ and the Average mAP (AmAP) over ten thresholds of $[0.05:0.95]$ with step at $0.05$. 
Due to the strong performance and publicly availability of its code, we adopt G-TAD~\cite{xu2020g} as method for TAL.
We followed its training and testing protocols without any modifications on ActivityNet-1.3.
The original G-TAD uses a RGB and optical-flow pre-trained TSN~\cite{wang2016temporal} video encoder.
We replace the video encoder of G-TAD with ours without the flow stream in favour of simplicity and efficiency.

We compare CLAP with other pre-training methods that use the same backbone and experimental setup based on G-TAD, such as, TAC~\cite{tran2018closer}, LoFi-TAL~\cite{LOFITAL}, TSP~\cite{alwassel_2021_tsp} and VideoClip~\cite{xu2021videoclip}.
We experimented with several variants of CLAP. The variants indicated with parenthesis (CLAP$^{(\cdot)}$) only employ synthetically-generated language prompts, but not dense captions.
\emph{CLAP$^{(clip)}$} where we use the synthetically generated prompts for training and include all the background clips in the loss as in Eq.~\eqref{eq:clip_loss},
\emph{CLAP$^{(mask)}$} where we use the language templates for training and the masked version of our loss as in Eq.~\eqref{eq:proposed_loss},
\emph{CLAP} where we use both the language templates and caption annotations and the masked loss, and
\emph{CLAP$^{no-cls}$} that omits the classification loss from \emph{CLAP} .

\paragraph{Results.} 
Table~\ref{tab:TAL_main} reports the performance of all compared methods and variants of our proposed video encoder.
First of all, we can observe that our baseline model CLAP$^{(clip)}$ already outperforms other pre-training competitors by $0.89\%$ (TAC), $0.05\%$ (TSP) in terms of AmAP, and $2.16\%$ (TAC), $1.09\%$ (TSP) in mAP@0.5.
This indicates that simple prompt engineering of templates in combination to the contrastive loss function is beneficial even for the TAL task, for which language is not that relevant.
When adding the proposed masked loss function (CLAP$^{(mask)}$), we obtain an improvement of $0.08\%$ AmAP.
We obtain the best results with CLAP, which also considers descriptive language during training, with a further improvement of $0.39\%$ AmAP ($+0.52\%$ compared to TSP).
This shows that using language templates in combination with captions leads to TAL accuracy.
Moreover, the best TSP variant with R(2+1)D--18 backbone performs worse than CLAP by $1.19\%$ mAP@0.5.
When we compare {CLAP$^{no-cls}$} and CLAP, the AmAP drops by approximately $1\%$ showing the importance of the classification loss.
Finally, we also see that the CLAP maintains its superiority over TSP on the Res3D--18 backbone (AmAP of $34.65\%$ vs. $34.10\%$).

\paragraph{Comparison to VideoCLIP~\cite{xu2021videoclip}.} 
There are 3 fundamental differences between VideoCLIP and CLAP.
First, VideoCLIP is trained on HowTo100M~\cite{miech19howto100m}, which is a large-scale dataset with 134k hours (1.2M videos) while CLAP uses ActivityNet which contains 849 hours (20k videos).
Second, VideoCLIP uses S3D~\cite{S3D} as video encoder that has higher FLOPs than R(2+1)D-18 used by CLAP (approx. $41$ vs $71$ GFLOPs).
Third, VideoClip employs transformer networks to learn interactions between multiple video clips, while CLAP relies on a simple combination of layers and contrastive loss.
Despite these differences, all variants of our method perform significantly better than VideoCLIP (Table~\ref{tab:TAL_main}).
This further supports our claim that CLAP representation is tailored for temporal localization by contrasting foreground and background of activities in untrimmed videos, while VideoCLIP provides more generic features.

\paragraph{Robustness study.} 
We compute the standard deviations of all the metrics from 5 runs with different initial states to study the effect of randomness in our model. 
From the last row of Table~\ref{tab:TAL_main}, the variance of mAP at a single tIoU threshold is around $0.1\%$ to $0.2\%$, and the standard deviation of average mAP is only $0.07\%$. 
Since our performance gain is $0.52\%$ AmAP when compared with TSP, our improvement on TAL is evident.

\paragraph{Feature quality.} 
In addition to our quantitative experimental comparison for the TAL task, we perform an analysis of the quality of video features extracted using either the pre-trained TSP~\cite{alwassel_2021_tsp} or our method. 
Figure~\ref{fig:dist_hist} illustrates the histogram of difference value between two $L2$ distances: foreground-to-foreground (\emph{fg2fg}) and foreground-to-background (\emph{fg2bg}). 
From the validation set of ActivityNet, for each video we select random two foreground clips of length one second and one background clip from the same video. 
Then, we extracted the features for these clips with the pre-trained video encoder and then take the $L2$ difference between the two foreground features (\emph{fg2fg}) and one of the pairs of foreground and background (\emph{fg2bg}). 
We subtract the distances (\emph{fg2bg} - \emph{fg2fg}). Hence, positive values are desired. 
As it can be observed from the figure, the distribution of our distance differences has a long tail centering around $~1.0$ in the x-axis (region B in Fig.~\ref{fig:dist_hist}), indicating that the video encoders pre-trained by our approach is more distinguishable for a significant part of the data.

\subsection{Generalization to Unseen Categories}
\label{sec:few_shot_TAL_res}
To demonstrate the generalization capabilities on unseen categories of the proposed method, we experimented in a few-shot learning setup.
Few-shot temporal action localization aims to adapt a model to unseen classes represented by as few as a single video.
We measure the performance in terms of tIoU 
as TAL setup.

For all pre-training methods, we adopted QAT~\cite{nag2021fewshot} as method for few-shot TAL since it works particularly well on the few-shot regime.
We replace the video encoder of QAT with the pre-trained ones and follow the 5-shot experimental setup recently introduced in~\cite{nag2021fewshot}. 
In particular, 200 action categories of ActivityNet-v1.3 are divided into train, val and test sets with ratios of 90\% (160 classes), 10\% (20 classes) and 10\% (20 classes). 
Hence, 20\% of the categories (in val and test sets) remains unseen during the pre-training stage for all our variants and baselines.



\input{tables/FewShotTAL_and_CaptionGrounding.tex}

\paragraph{Results.} 
We compare CLAP with recent pre-training methods that use the same backbone and experimental setup, such as QAT~\cite{nag2021fewshot}, TAC~\cite{tran2018closer}, TSP~\cite{alwassel_2021_tsp}, and additionally with features extracted from VideoCLIP~\cite{xu2021videoclip} model that was trained on Howto100M dataset.
Table~\ref{tab:TAL_fewshot} shows the benefits of adopting our masked loss and using language during pre-training.
In fact, CLAP$^{(clip)}$, which only uses language templates, performs better than TSP in terms of AmAP ($39.07\%$ vs $38.70\%$).
We obtain improvement  with the masked loss, especially in mAP@0.5.
And we obtain the best results of $40.25\%$ in average mAP when additionally using captions during training.
It is interesting that language becomes more important for few-shot TAL rather than in supervised setup because the free-form language carries information about unseen activities.
This can also be supported with the performance of VideoCLIP~\cite{xu2021videoclip} that was trained on significantly larger scale dataset (Howto100M: $134k$ vs. Activity-Net: $.8k$ hours) video data with pairs of video and transcribed audio as VideoCLIP~\cite{xu2021videoclip} achieves +$.42\%$ AmAP than TSP. On the other hand, CLAP performs better than VideoCLIP by $1.13\%$, showing the efficacy of using the contrastive loss in a smaller dataset.

\begin{figure*}[t!]
    \centering
    \begin{subfigure}[b]{0.55\textwidth}
\includegraphics[width=0.99\linewidth]{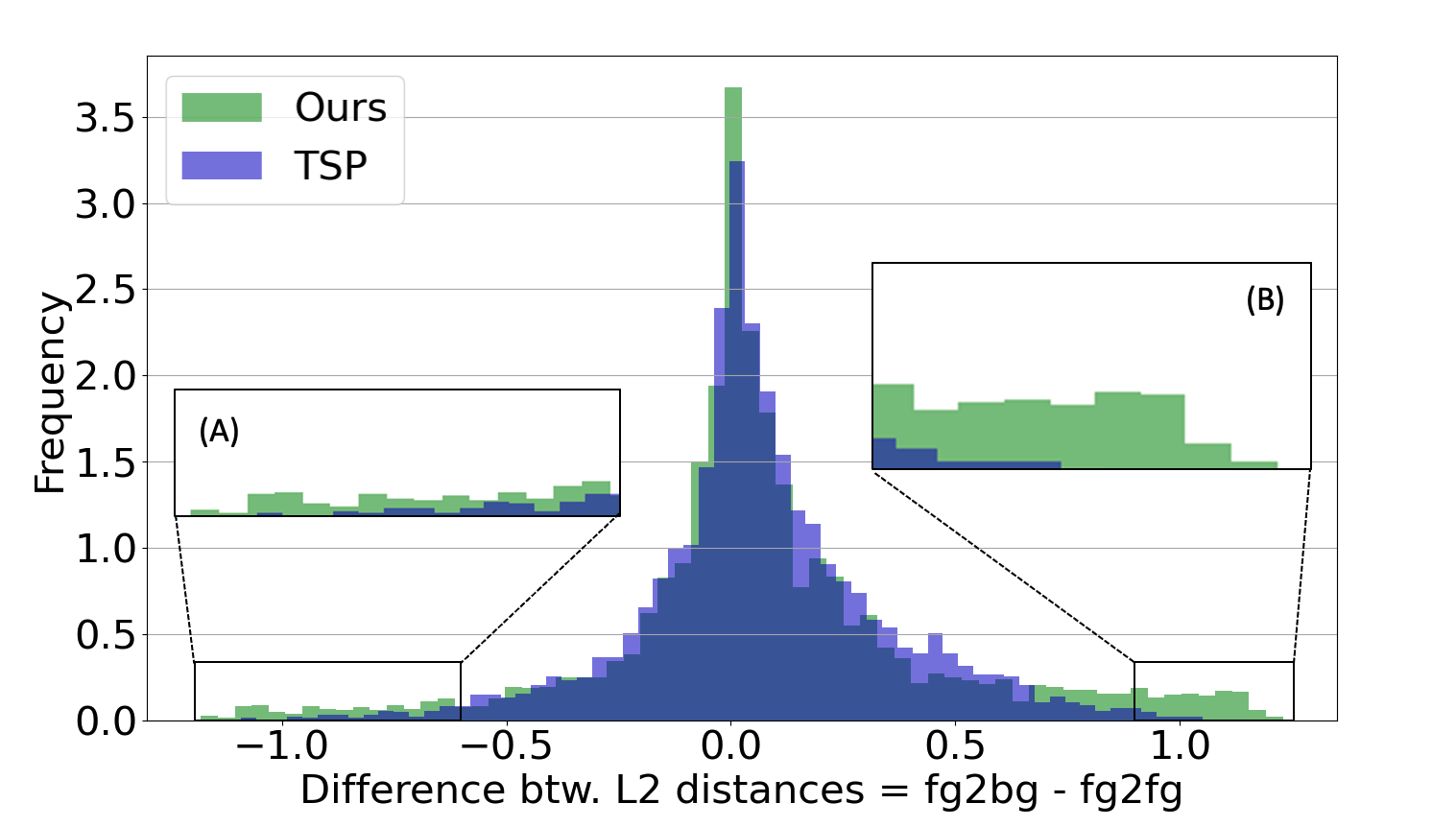}
\caption{}
\label{fig:dist_hist}
    \end{subfigure}%
    ~
    \begin{subfigure}[b]{0.45\textwidth}
        \centering
\includegraphics[width=0.99\linewidth]{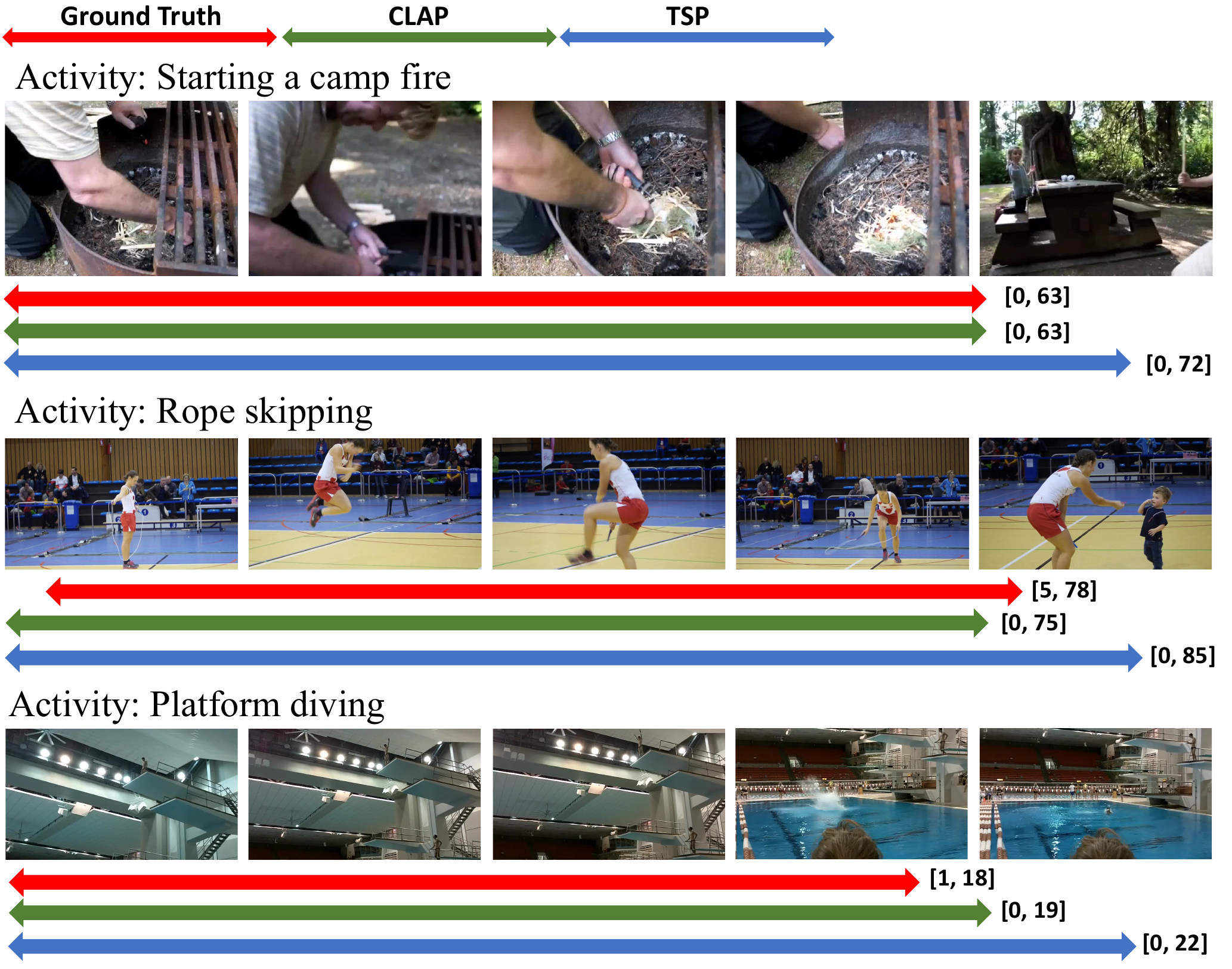}
\caption{}
\label{fig:qualitative}
    \end{subfigure}
\caption{
\textbf{(a) Feature similarity.} We compare our features and those of TSP~\cite{alwassel_2021_tsp}. \emph{fg2fg} stands for the distance between features of foreground and \emph{fg2bg} is the one between features of foreground and background within the same video. \emph{c.f.} Section~\ref{sec:TAL_res} for details.
\textbf{(b) Qualitative comparison of CLAP and TSP in few-shot TAL setting.} The arrows under each frame sequence indicate the start and end time of the predicted or ground truth actions, which are also shown in square brackets as $[start, end]$ in seconds.
}
\end{figure*}

\paragraph{Qualitative results.}
Fig.~\ref{fig:qualitative} presents a qualitative comparison for CLAP and TSP between predictions and ground-truth intervals of the actions. 
One can notice that TSP tends to predict more frames beyond the actual foreground interval. 
On the other hand, CLAP could discard the frames although they are relevant to the activity. 
For instance, a wooden bench in the end of the \emph{starting a camp fire} has a semantic relation to camping but not the activity. 
Similarly, when a kid enters the scene in the \emph{rope skipping}, CLAP can make an early cut in the prediction while TSP did not, 
presumably CLAP can learn additional semantic relations of activities and presence of a child. 
Such examples are important indications that our model learns to generalize from a base dataset with the proposed approach. 

\subsection{Generalization to Language-Video Tasks}
\label{sec:temporal_video_caption_grounding}

We investigate the task of video language grounding which has the
objective to localize the segment in the video that best matches a given free form language query. 
This is a more challenging task than TAL because the query in the test set is usually never seen during training. 
In addition, we designed the experiment to deal with a difference in the data distribution between pre-training and fine-tuning sets.
We use the Charades-STA and TACoS for testing, while pre-training on ActivityNet-1.3.
This experiment analyzes the generalization capability of pre-trained video encoders.
We adopt 2D-TAN~\cite{2DTAN_2020_AAAI} as the method for this downstream task due to its promising performance and publicly available code. 
Similarly to our previous setup in TAL tasks, we replace the video encoder with the original one where they adopted VGG features~\cite{simonyan2014very}. 
We use top-1 recall at two $tIoU$ thresholds, i.e., $\{0.5, 0.7\}$ and the mean IoU of each method.

\paragraph{Results.} 
We compare CLAP and our variants with other pre-training methods that use the same backbone and experimental setup, such as, 2D-TAN~\cite{2DTAN_2020_AAAI}, TAC and TSP.
As shown in Table~\ref{tab:caption_grounding}, we notice a significantly higher performance compared to 2D-TAN (\emph{e.g.} $42.8$ vs.\ $49.24$ recall for IoU@$0.5$) due to the difference in how the video clips are represented.
On the other hand, all other variants outperform TSP by a significant amount (all of them $>2\%$).
This shows that the video encoder pre-trained on a different dataset can achieve state-of-the-art results, and the proposed loss is important to achieve the best performance.

\paragraph{Synthetic prompt vs.\ video caption.} 
Unlike in the results for TAL, using textual description is not always better than synthetic prompt. 
For example, CLAP is $0.19\%$ worse than CLAP$^{(mask)}$ on the metric IoU@0.5. We assume that this is because of the domain gap between textual descriptions in ActivityNet and Charades. ActivityNet Caption contains mostly specific languages to describe the actions such as ``A person is drinking from a coffee cup and looking away from the camera", while Charades' descriptions are more concise, \textit{e.g.} ``person opens the door". Since our prompt description is more simple and generic, the corresponding model can be well-transferred to Charades data. 
To verify our assumption, we conducted a CLAP$^\dagger$ experiment with prompt and caption language for foreground clips, and only prompt language for background clips. 
As shown in the last row of Tab.~\ref{tab:caption_grounding}, CLAP$^\dagger$ outperforms both CLAP and CLAP$^{(mask)}$ on mean IoU with a slight margin.
CLAP variants outperform compared methods by a large margin for IoU@0.5.

\input{tables/VLG_Tacos_main.tex}
Similar to the experiments on Charades-STA, we compare CLAP on TACoS with other pre-training techniques in Tab.~\ref{tab:tacos} using 2D-TAN as the main method. 
We measure performance in terms of recall@1 at IoU thresholds and mean IoU.
It is worth noting that CLAP performs better than TSP and 2D-TAN on most of the metrics. 
Moreover, using language during pre-training (4$^{th}$ row) can make 2D-TAN predict more precise results and produce higher recall on high IoU thresholds of 0.5 and 0.7.
We believe this is because the CLAP pre-trained video model is more aware of language and have a stronger correlation to actions. We also tried CLAP$^\dagger$ on TaCoS and have not seen a significant performance difference, which can be attributed to the longer language descriptions compared to Charades-STA ($6.23$ vs. $9$ words per description).


%% file: tables/TAL_main.tex
\begin{table*}[t]
\centering
\small
\caption{\textbf{Results for TAL on ActivityNet-1.3 using G-TAD \cite{xu2020g}.} 
We report the mAP at different tIoU thresholds and the average mAP.
We denote with CLAP$^{(\cdot)}$ the method which uses only language templates. TR stands for Transformer network in VideoCLIP~\cite{xu2021videoclip}. Please refer to Sec.~\ref{sec:TAL_res} for details.}
\label{tab:TAL_main}
\resizebox{0.85\linewidth}{!}{    
\begin{tabular}{l|c|c|c|c|c|c }
\hline    
       & \textbf{Pre-train} & \textbf{Video} &   \multicolumn{3}{c|}{\textbf{mAP, tIoU@X } }& \textbf{Avg.} \\ 
 & \textbf{Dataset} & \textbf{Backbone} & \textbf{0.5} & \textbf{0.75} & \textbf{0.95} & \textbf{mAP} \\ \hline
TAC~\cite{tran2018closer}                                                                               & K400                         & R(2+1)D-18              & 47.57                 & 33.11                  & 8.10                    & 32.46             \\ \hline
TAC~\cite{tran2018closer}                                                                               & ANET                         & R(2+1)D-18              & 49.00                    & 34.56                  & \textbf{9.42}                   & 33.87             \\ 
LoFi-TAL~\cite{LOFITAL}                                                                          & ANET                         & R(2+1)D-18              & 49.84                 & 34.73                  & 8.64                   & 34.21             \\ 
TSP~\cite{alwassel_2021_tsp}                                                                       & ANET                         & R(2+1)D-18              & 50.07                 & 35.61                  & 8.96                   & 34.71             \\
TSP~\cite{alwassel_2021_tsp}                                                                       & ANET                         & Res3D-18              & 49.81                 & 34.81                  & 8.63                   & 34.10             \\
VideoCLIP~\cite{xu2021videoclip}                                                                       & HowTo100M                         & S3D+TR              & 48.95                 & 34.59                  & 8.15                   & 33.73             
\\ \hline 
CLAP$^{(clip)}$ & ANET                         & R(2+1)D-18              & 51.16                 & 35.44                  & 8.65 & 34.76           \\
CLAP$^{(mask)}$ & ANET                         & R(2+1)D-18              & 51.24                 & 35.30                   & 8.96                   & 34.84             \\
CLAP$^{no-cls}$ & ANET                         & R(2+1)D-18              & 50.36                 & 34.57                   & 8.91                   & 34.18             \\ \hline
CLAP                                                                  & ANET                         & Res3D-18              & {50.92}                     & {35.35} & 8.59                  & {34.65} \\ 
CLAP                                                                  & ANET                         & R(2+1)D-18              & \textbf{51.55}$\pm$ 0.10                     & \textbf{35.80}$\pm$ 0.17 & {9.26}                     $\pm$ 0.23 & \textbf{35.23} $\pm$ 0.07\\ 
\hline 
\end{tabular}
}
\end{table*}

%% file: tables/FewShotTAL_and_CaptionGrounding.tex
\begin{table}
\caption{We denote the method which uses only language templates with CLAP$^{(\cdot)}$. CLAP$^{\dagger}$ uses background with synthetic prompts and action with both synthetic prompts and captions. The backbone of VideoCLIP is S3D~\cite{S3D} and the transformer head while the rest has R(2+1)D. 
\textbf{(a) Results for few-shot TAL on ActivityNet-1.3} using QAT~\cite{nag2021fewshot} with pre-training strategies.
\textbf{(b) Results for language grounding on Charades-STA} using \cite{2DTAN_2020_AAAI} and different pre-training strategies.
}
\begin{subtable}[c]{0.49\textwidth}
\centering
\subcaption{}
\label{tab:TAL_fewshot}
\centering
\resizebox{0.98\linewidth}{!}{  
\begin{tabular}{l|c|c|c|c|c}
\hline    
       & \textbf{Pre-train} &   \multicolumn{3}{c|}{\textbf{mAP, tIoU@X } }& \textbf{Avg.} \\ 
      & \textbf{Dataset} & \textbf{0.5} & \textbf{0.75} & \textbf{0.95} & \textbf{mAP} \\ \hline
QAT~\cite{nag2021fewshot}                     & K400                         & 65.25                 & 37.60                   & 3.74                   & 37.37             \\ \hline
TAC~\cite{tran2018closer}& ANET-160                     & 65.94                 & 38.61                  & 3.31                   & 38.06             \\ 
TSP~\cite{alwassel_2021_tsp} & ANET-160                     & \textbf{67.73}                 & 38.62                  & 4.71                   & 38.70              \\ \hline 
VideoCLIP~\cite{xu2021videoclip} & ANET-160                     & 66.11                 & 38.16                  & 7.83                   & 39.12              \\ \hline 
CLAP$^{(clip)}$ & ANET-160                     & 65.75                 & 38.93                  & \textbf{6.85}                   & 39.07             \\ 
CLAP$^{(mask)}$            & ANET-160                     & 67.12                     & 39.59                      & 6.00                      & 39.32                 \\ 
CLAP         & ANET-160                     & {67.58                   } & \textbf{39.77}                      & 6.62                      & \textbf{40.25}            \\ \hline    
\end{tabular}}
\end{subtable}
\begin{subtable}[c]{0.49\textwidth}
\centering
\subcaption{}
\label{tab:caption_grounding}
\centering
\resizebox{0.98\linewidth}{!}{
\begin{tabular}{l|c|c|c}
\hline
 &   \textbf{IoU@0.5} & \textbf{IoU@0.7} & \textbf{mIoU} \\ \hline
2D-TAN~\cite{2DTAN_2020_AAAI}                   & 42.80                 & 23.25                 & N.A.     \\ \hline
TAC~\cite{tran2018closer}                     & 46.57                & 26.88                  & 41.46     \\ 
TSP~\cite{alwassel_2021_tsp}                      & 46.27                & 26.01                & 41.54     \\ \hline
CLAP$^{(clip)}$ & 49.00                & \textbf{28.71}        & 43.62 \\ 
CLAP$^{(mask)}$             & \textbf{49.43}                    & 28.20       & 43.70                     \\ 
CLAP          & 49.24                & 28.12               & 43.36    \\   \hline    
CLAP$^\dagger$          & 49.22                & {28.34}   & \textbf{43.94}   \\  
\hline
\end{tabular}}
\end{subtable}
\end{table}

%% file: tables/VLG_Tacos_main.tex
\begin{table}[!ht]
\centering
\scriptsize
\caption{
{Results for video language grounding on TACoS using 2D-TAN \cite{2DTAN_2020_AAAI} and comparing different pre-training strategies.}
}
\resizebox{0.8\linewidth}{!}{ 
\begin{tabular}{l|c|c|c|c|c}
\hline
 & \textbf{IoU@0.1} & \textbf{IoU@0.3} & \textbf{IoU@0.5} & \textbf{IoU@0.7} & \textbf{mIoU} \\ \hline
2D-TAN~\cite{2DTAN_2020_AAAI} & 47.59 & 37.29 & \textbf{25.32} & N.A. & N.A. \\ 
TSP~\cite{alwassel_2021_tsp} & 51.71 & 37.37 & 22.82 & 10.30 & 24.91 \\ \hline
CLAP$^{(mask)}$ & \textbf{53.14} & \textbf{38.02} & 22.77 & 9.50 & 25.18 \\  
CLAP & 52.81 & 37.47& 24.69 &\textbf{10.97}& \textbf{25.70} \\ \hline 
\end{tabular}}
\label{tab:tacos}
\end{table} 

%% file: sections/5_conclusion.tex
\section{Conclusion}
\label{sec:conclusion}

We present CLAP, a novel contrastive language-action post-pre-training method for temporal localization.
CLAP bridges the gap between the universal pre-training and fine-tuning for downstream tasks by aligning the video representation with textual descriptions and the capability of correlating video clips to leverage their relationship.
We show that CLAP significantly improves state-of-the-art methods on the temporal action localization, and it also has a good generalization for unseen categories in a few-shot learning setup and video language grounding task.